# Single-Pass, Adaptive Natural Language Filtering: Measuring Value in User Generated Comments on Large-Scale, Social Media News Forums

By

Manuel Amunategui

Thesis Project

Submitted in partial fulfillment of the

Requirements for the degree of

Master of Science in Predictive Analytics

Northwestern University

December 2016

Professor Kenneth H Stehlik-Barry, First Reader

Professor Edward Arroyo, Second Reader

# Table of Contents

## Abstract

There are large amounts of insight and social discovery potential in mining crowd-sourced comments left on popular news forums like Reddit.com, Tumblr.com, Facebook.com and Hacker News. Unfortunately, due the overwhelming amount of participation with its varying quality of commentary, extracting value out of such data isn't always obvious nor timely. By designing efficient, single-pass and adaptive natural language filters to quickly prune spam, noise, copy-cats, marketing diversions, and out-of-context posts, we can remove over a third of entries and return the comments with a higher probability of relatedness to the original article in question. The approach presented here uses an adaptive, two-step filtering process. It first leverages the original article posted in the thread as a starting corpus to parse comments by matching intersecting words and term-ratio balance per sentence then grows the corpus by adding new words harvested from high-matching comments to increase filtering accuracy over time.

## Introduction

This paper presents an adaptive filtering approach to measure the degree at which crowd-driven commentary is related to an originating article. The method matches comments against the original article based on the quantity of intersecting words and the balance of words found in each sentence. It also keeps track of new words from high-scoring comments and injects them into the original list of accepted words.

This approach offers many advantages over traditional sentence matching as it gets more accurate as it processes more comments, it adapts to long threads by allowing a certain amount of drift from the original topic, and requires no pre-processing. In essence, it builds an evolving, quantitative profile of an article and the crowd-sources discussion around it.

In this paper, the accuracy of this method is measured using 'karma points'. This is a member scoring system used by not only Hacker News but other sites such as Reddit.com, that rewards good participants with a score derived from the number of up and down votes that he or she has received. The higher the Karma points a member has, the better his or her standing in that online community. As the goal of this filtering algorithm is to find relevant content, the karma score will be used as an objective proxy for relevance.

Due to the simplicity of this filtering pipeline, this approach should easily scale to real-time, high-throughput social websites such at Twitter or Facebook.

## Justification

Determining if a comment is on topic is a challenging task. If the decision is based on a large corpus, such as the Internet (Preslav Nakov, 2008) or labor intensive, human curated and scored lists (J. L. McClelland, 1987), then relatedness can be learned and trained by methods such as bag-of-words, vector representation or classification. On the other hand, when the corpus is limited between 500 to 2000 words, the range found in a typical news article, the common natural language processing approach tends to be rule-based sematic filters (Radu Vlas, 2011). Designing rule-based systems for message groups and social media web sites where topics are not known in advance and where the culture and style changes constantly isn't feasible in determining whether a comment is on topic or not. Instead, combining multiple approaches may be a better solution to quickly capture the essence of the subject at hand and optimize the filtering over time by allowing it to keep learning and get more accurate as it processes more comments.

## Review of Literature

Natural language processing (NLP), topic modeling, and sentence similarity processing are all closely related to the topic at hand. The challenge with the presented approach is having the ability to quickly recognize the topic at hand, keep processing time to a minimum, be adaptive to drifting conversations, while filtering real-time, large data streams.

## Bag-of-Words (BOW)

One of the most popular NLP method after the classic rule-based approach is the bag-of-words. The term was coined by Zellig Harris (Bruce E. Nevin, 2002) in the 1950's and the concept is simple and heavily relied upon in the comment filtering pipeline presented here. A bag-of-words is nothing more than a list of words, deemed of importance, that you can use to measure relatedness against other bodies of text. You can easily quantify that relatedness by counting the words that intersect a body of text with those in the bag. BOW has many additional advantages such as being highly customizable. It can contain all the words in the English language to filter out foreign languages or contain a small subset to focus on specialized vocabularies such as medical, technical, urban, historical, fictional terms, etc.

Hyman (Hyman, 2010) demonstrates how a simple method for document retrieval using BOW and word weights can yield equivalent recall compared to more complex methods such as topic classification or content indexing.

The approach presented in this paper also uses BOW along with weights derived from the amount of times those words are used. This allows matching content not only using words but also using frequency of use. This ensures that filters capture typical comments while avoiding edge cases even if they both contain words found in the BOW. Viewed on a weight spectrum, words with low frequency, tend to contain rare, misspelled or useless tags, those with high frequency contain overly used words such as stop words, while the middle contains a balanced mixture of words which should have a higher probability of being representative of the topic at hand.

Term Frequencies and Term Frequency–Inverse Document Frequency (TFIDF)

Another popular approach is the use of term frequencies and frequency-inverse document frequency. TFIDF gathers terms found in a document and calculates the amount of times they appear. It can then use these term frequencies to compare other documents. It uses the log of the terms found to not give too much weight to rare or to heavily used terms. This dampening effect is also applied in this comment filtering pipeline but in a much simpler manner, by calculating the article's term frequencies on a sentence basis.

Juan Ramos (Ramos, 2013) shows how TFIDF can be used to accurately retrieve documents by using simple, user-entered queries on small sets of documents. He demonstrates how such approach offers a fast and transparent system for document retrieval. He also points out some the weaknesses of this approach, such as not taking in consideration the relationship between words nor the ability of matching similar words. As the collection of documents gets larger and more variations of similar words are collected, the precision decreases while the processing time increases.

Latent-Semantic Indexing (LSI)

Latent-Semantic Indexing uses statistical calculations to create relationship concepts that it can use to categorize new document within topics. Unlike TF-IDF, it can find similarities between documents and topics without needing overlapping terms. Chen, Martin, Daimon, Maudsley (Hongyu Chen, 2013) demonstrate the power of LSI in the biological and biomedical knowledge arena. By being able to retrieve information from an ever growing corpus, and more interestingly, find links between published articles that would not have been possible with rule-based methods or Boolean searches, makes this a powerful and surprising tool.

Xi-Quan Yang (Xi-Quan Yang, 2009) illustrates the development of specialized ontologies to work as glue between large topics with little intersecting information but similar themes. This is one of the disadvantages of LSI as it requires carrying dictionaries of concepts and parsing through them to find links between disparate documents.

This drawback is addressed in the presented comment filtering pipeline by allowing the filter to adapt over time. Instead of assuming a large ontology from the start, the original article itself is the ontology and each comment has the ability of growing it. This adaptive filtering pipeline offers a much more economical way of finding similarities without cost of creating numerous dictionaries in hope of covering all posted articles.

Sentence Similarity

Sentence similarity is an integral approach to the comment filtering pipeline presented in this paper. Instead of only looking at intersecting words between the original article and comments, we also calculate a weight of word usage for each sentence by dividing the number of intersecting words in each sentence by the total number of words in the original article. This yields a single benchmark value for the article and comparative values for each comment. This approach also gives precedence to full sentences, longer comments, while penalizing partials or single words.

Many papers are exploring more complex approaches to the challenging task of matching short sentences. Li, McLean, Bandar, O'Shea, Crockett (Y. Li, 2006) attempt to solve this problem by checking word order and word-pairs in different contexts. Their approach relies on using large collections of word pairs culled from the Internet in order to assign meaning to new sentences with these matching word markers. Unfortunately, ontology-based approaches aren't appropriate for use in fast data streams as they would prove slow, clunky and require enormous topic ontologies.

## Semantic Relations

Another popular method is to find the relationship between nouns using verbs in sentences to understand some of the context within the sentence. Nakov and Hearst (Preslav Nakov, 2008) showed that by leveraging the relationships between nouns and verbs they could understand the sentence's context and relationship between two nouns. Once again, their approach requires a large corpus of relationships gathered from the web or other sources and therefore may not be appropriate for the filtering project at hand due to performance and/or topic expertise needed with complex ontologies.

## Word-representation Vectors

A powerful way of measuring the relationship between words in a large corpus is through the use of word-representation vectors. Mikolov, Yih, Zwei (Tomas Mikolow, 2013) in their ground-breaking paper on word representations showed that mathematical relationships can be found in large corpuses. They use as example the querying of a relationship based on 'woman' and 'king' but not 'man':

$$(king + woman) - man \rightarrow queen$$

The model yields the word 'queen'. They demonstrate that a lot of relationships in sentences are quantitative in nature like masculine and feminine, singular and plural, different topic words, etc. The drawbacks of this method are the necessity of running a narrow neural network on the corpus in order to map these relationships which is time consuming. It has also shown to work best with extremely large bodies of text. Obviously, the speed requirements and the size of the articles and comments disqualify this approach for the comment filtering pipeline.

## Methods

This study uses data from Hacker News (HN), the crowd-sourced site created by Y Combinator. HN users post a constant flow of news, current events, and general-interest articles via pasting the actual content or just the URL link. Site participants liberally comment on the posted content as well as commenting on each others comments. Hacker News is transparent about its data and offers simple ways of accessing it without complicated rules or authentication schemes. HN serves as a good proxy for other crowd-sourced news sites that don't offer such easy access or where the data streams are too voluminous therefore requiring much more complex capturing mechanism.

A Python script (see appendix) is hard-coded to collect the comments from our different threads. By using different articles containing different quantities of comments, we can better determine how and where this approach does well and where it struggles. The corresponding articles for these 10 comment threads are downloaded manually.

A second Python script (see appendix) runs the model by first parsing the article and subsequently the comment threads. The final output of the script is a summary of the total amount of comments, the amount of comments filtered by using the first filter alone and using both filters. HN also scores its members with a formula calculated using the amount of commentary activity, and whether they are up-votes or down-votes. They call this 'Karma' points. This study will rely on this scoring system as a quantitative benchmark to measure the total mean karma points from comments filtered out versus those kept. Comment samples from both the accepted and rejected groups will also be extracted for visual validation.

## Filtering and Adaptive Filtering Algorithms

Traditional NLP approaches are not designed to handle big data or high-speed data streams. Stemming, stop-words, inverse term-frequencies, string distance, deep learning classification etc., are eschewed for simple numerical comparison using single-pass profiling and an adaptive algorithm.

The first algorithm parses the original article and builds a word map of topic terms then calculates the average use per sentence. The same map is used on incoming comments to determine if they are on-topic and by how much. This approach yields a single quantitative value of relatedness. If the comment's value is within an acceptable threshold of the original article, the comment is deemed on-topic and accepted.

The adaptive algorithm monitors comments with a high-filter score of relatedness and adds words not found in the master set of topic words. This allows the filtering mechanism to grow its list of acceptable words and broadens its understanding of what is topical. It also allows the conversation to account for some progressive topic drift.   In essence, the first comments will have to intersect with the words from original article, but as the conversation progresses, the filtering algorithm will be able to handle growing degrees of separation from the original article and still maintain the overall topic. The quantity of drift can be control by either lowering or raising the adaptive filter's threshold. A higher threshold will be more conservative in letting new words come in, therefore reducing topic drift.

## Initial Natural Language Filtering Algorithm

The crux of the first filtering algorithm is to build the topic vocabulary list and its usage weight in sentences. The original article represents the ground-truth content. Getting a clear quantitative grasp of the content is critical as it will immediately affect the triage of subsequent comments.

To maintain a certain degree of intersection between words in both the article itself and the comments, the text is stripped of special characters, numbers, short words, and forced to lower case.

Three objects are needed to quantitatively filter comments using the original article:

- A word-list containing all 'cleaned' word occurrences
- A list of 'cleaned' sentences
- The number of sentences in the article

The article is cleaned by removing non-alphabetic characters and words of unacceptable length (with the exception of acronyms that are automatically kept by keeping short capitalized sets of words), the article is split by words to make up the word list and by sentences to make up the sentence list.

The article's word list is used to create the initial set of topic words. The number of words in each cleaned sentences is then divided by the number of topic words. The mean of the resulting values of all sentences is then multiplied by the number of sentences in the original article:

$$ArticleScore = \mu(\frac{SentenceWordCount}{Master\ ListWordCount}) * ArticleSentenceCount$$

Using the above formula, we hold a single quantitative representation of the originating article that takes into consideration individual sentence word ratio and overall article length. With this benchmark score, the filter can start processing incoming comments.

Calculating a Comment's Relevancy

The comments are also stripped of numerical and special characters and forced to lower case. We then calculate the term-ratio score per sentence by dividing the count of intersecting words by the master topic word list count. We then multiply the mean of all ratios by the number of sentences in the comment, just like we did with the original article, to get our final quantitative score for the comment.

$$CommentScore = \mu(\frac{IntersectWordsInSentenceCount}{MasterListWordCount})*CommentSentenceCount$$

The value is then compared against the article's score and if it is at or above the user set threshold, it is considered acceptable and the comment is deemed on-topic.

### Finding a Permissible Threshold

The threshold value is manually selected as it will depend on a reader's tolerance and needs. Comments will have a lower score than the benchmark so we need to account for those lower scores. Trial and error is one way to find what percentage of the benchmark score is acceptable. If a reader wants more relevancy and is okay with fewer comments, then the threshold can be raised. On the other hand, if they want more comments and only need to prune the really bad ones, then the threshold can be lowered.

### Reasoning Behind First Algorithm

Cleaning the words works well for short articles as it allows more terms to overlap and yield a more interesting quantitative weight value when data matches. On the other hand, if larger documents, books or bodies of knowledge were used instead of articles, leaving the words exactly as they are read would be preferable as it would entail less transformation and processing time and a closer quantitative grasp over the content – a word that is capitalized shouldn't necessarily be considered the same as one that isn't. On the flip side, If the articles are consistently very short, then stemming each word may be necessary.

Regarding term frequencies, the most common terms are usually prepositions or articles, while the less frequent terms are usually misspellings, links, non-word sets, etc. Here an assumption can be made to trim the top most frequent terms and the bottom least frequent ones. A lot of the most frequent ones are trimmed out by cutting any word that is less than three characters. The rare ones don't pose much of a threat as they will almost never find matches amongst comments but there could be processing speed advantage in removing them from the lists used during filtering – if they are too large, they may affect filtering speeds.

Adaptive Natural Language Filtering Algorithm

The second algorithm is used to add new words to the original frequency list beyond those found in the article. These new words are harvested from comments that are confirmed very relevant to the article, according to a more demanding threshold, if the word(s) isn't already in the list, it is added.

The more comments this second algorithm gets to process, the larger its list of acceptable words grows, and the better its understanding and filtering capabilities become. Given enough filtering time, it is possible that this second algorithm can accept comments with zero words intersecting between the comment and the original article.

The adaptive filter has its own threshold setting. It should be higher than the base filter to insure that it only adds relevant words to the topic word list.

### When to Add Missing Comment Words to Master Topic Word List

A key aspect to the second algorithm is when to add or not to add new words. When a commentator gets a high score for the overall comment, the pipeline can assume that this is a quality comment, so if there are words missing in the master topic list, we can add them. This also is done via trial and error and may vary by content type. A good rule of thumb is to make it a certain percentage higher than the base filter threshold.

### Reasoning Behind Second Algorithm

The first algorithm is capable of running on its own without the addition of the second adaptive filter. As a matter of fact, that is exactly how the comment filtering pipeline needs to work at the beginning of a thread. If the threshold is kept down, then a larger amount of comments will get through but will degrade the quality of the comment filtering. Adding the second algorithm attempts to let the filtering mechanism gradually better its understanding of the participants' communication patterns. The drawback of this approach is that the pipeline may reject more comments initially until it grows a better understanding of the topic.

## Validating the Comment Filtering Pipeline with Karma Points and Common Sense

An interesting way to validate this approach quantitatively is to measure the karma points of each user. According to Hacker News, karma points are:

> "Roughly, the number of upvotes on their stories and comments minus the number of downvotes. The numbers don't match up exactly, because some votes aren't counted to prevent abuse." (News, n.d.)

The Hacker News algorithm isn't made public to discourage members from gaming the system. One known failsafe against abuse is that they only grant down-voting power to users with at least 500 karma points.

By comparing the Karma points given to each user based on social grade, we should see that our comment filtering pipeline will favor those with high scores versus low ones. It turns out this is a great way to grade both filters. We will attach the karma score to each comment as it goes through the filtering pipeline and track the scores of the commentators that are filtered out versus those that aren't. We need to keep in mind that commentators with multiple comments for a particular article can very well have comments falling into both, rejected and accepted, buckets.

# Results

As the appropriateness of a comment is subjective, we leverage the Karma score from the commentator as a proxy to the value of the comment. Also, to offer a diverse look at the performance of this pipeline, the following 10 stories from Hacker News are used:

1. Reddit.com post: Warning: Microsoft Signature PC program now requires that you can't run Linux. Lenovo's recent Ultrabooks among affected systems (BaronHK, n.d.).
2. Kalzumeus newsletter: An A/B Testing Story (McKenzie, n.d.).
3. Science Bulletin: Researchers teleport particle of light six kilometres (Technology, n.d.).
4. BBC article: A 16-year-old British girl earns £48,000 helping Chinese people name their babies (Newsbeat, n.d.) .
5. The Bizarre Role Reversal of Apple and Microsoft (Backchannel, n.d.).
6. Of course smart homes are targets for hackers (Garrett, n.d.).
7. Soylent halts sales of its powder as customers keep getting sick (Dave, n.d.).
8. Google AI invents its own cryptographic algorithm (Anthony, n.d.).
9. General questions about the Airbnb Community Commitment (Airbnb, n.d.).
10. Cognitive bias cheat sheet (Benson, n.d.).

| Story 1: Warning Microsoft Signature PC program now requires that you cant run Linux | | | |
|---|---|---|---|
|  | No Filter | 1st Filter | All Filters |
| Approved Comments | 438 | 42 | 60 |
| Rejected Comments | 0 | 396 | 378 |
| Approved Average Karma Points | 3619 | 4248 | 3876 |
| Topic Word List Count | 0 | 336 | 766 |
| Total Word Count in Article | 1007 | | |

| Story 2: An AB Testing Story | | | |
|---|---|---|---|
|  | No Filter | 1st Filter | All Filters |
| Approved Comments | 47 | 24 | 25 |
| Rejected Comments | 0 | 23 | 22 |
| Approved Average Karma Points | 8531 | **13826** | **13294** |
| Topic Word List Count | 0 | 673 | 870 |
| Total Word Count in Article | 2187 | | |

| Story 3: Researchers teleport particle of light six kilometers | | | |
|---|---|---|---|
|  | No Filter | 1st Filter | All Filters |
| Approved Comments | 154 | 110 | 131 |
| Rejected Comments | 0 | 44 | 23 |
| Approved Average Karma Points | 2882 | 2521 | **3151** |
| Topic Word List Count | 0 | 303 | 1549 |
| Total Word Count in Article | 820 | | |

| Story 4: A 16-year-old British girl earns £48,000 helping Chinese people name their babies | | | |
|---|---|---|---|
|  | No Filter | 1st Filter | All Filters |
| Approved Comments | 199 | 150 | 178 |
| Rejected Comments | 0 | 49 | 21 |
| Approved Average Karma Points | 3666 | 4290 | 3929 |
| Topic Word List Count | 0 | 191 | 1698 |
| Total Word Count in Article | 501 | | |

| Story 5: The Bizarre Role Reversal of Apple and Microsoft | | | |
|---|---|---|---|
| | No Filter | 1st Filter | All Filters |
| Approved Comments | 176 | 59 | 107 |
| Rejected Comments | 0 | 117 | 69 |
| Approved Average Karma Points | 4809 | 2782 | 5769 |
| Topic Word List Count | 0 | 321 | 1537 |
| Total Word Count in Article | 742 | | |

| Story 6: Of course smart homes are targets for hackers | | | |
|---|---|---|---|
| | No Filter | 1st Filter | All Filters |
| Approved Comments | 59 | 26 | 40 |
| Rejected Comments | 0 | 33 | 19 |
| Approved Average Karma Points | 3687 | 3311 | 3607 |
| Topic Word List Count | 0 | 206 | 800 |
| Total Word Count in Article | 549 | | |

| Story 7: Soylent halts sales of its powder as customers keep getting sick | | | |
|---|---|---|---|
| | No Filter | 1st Filter | All Filters |
| Approved Comments | 1001 | 542 | 893 |
| Rejected Comments | 0 | 459 | 108 |
| Approved Average Karma Points | 3289 | 3019 | 3370 |
| Topic Word List Count | 0 | 161 | 5318 |
| Total Word Count in Article | 321 | | |

| Story 8: Google AI invents its own cryptographic algorithm | | | |
|---|---|---|---|
| | No Filter | 1st Filter | All Filters |
| Approved Comments | 162 | 46 | 60 |
| Rejected Comments | 0 | 116 | 102 |
| Approved Average Karma Points | 2801 | 4047 | 4229 |
| Topic Word List Count | 0 | 231 | 941 |
| Total Word Count in Article | 653 | | |

| Story 9: General questions about the Airbnb Community Commitment | | | |
|---|---|---|---|
|  | No Filter | 1st Filter | All Filters |
| Approved Comments | 160 | 45 | 126 |
| Rejected Comments | 0 | 115 | 34 |
| Approved Average Karma Points | 14671 | 19049 | 15173 |
| Topic Word List Count | 0 | 109 | 1480 |
| Total Word Count in Article | 316 | | |

| Story 10: Cognitive bias cheat sheet | | | |
|---|---|---|---|
|  | No Filter | 1st Filter | All Filters |
| Approved Comments | 139 | 78 | 87 |
| Rejected Comments | 0 | 61 | 52 |
| Approved Average Karma Points | 1971 | 1946 | **2079** |
| Topic Word List Count | 0 | 862 | 1331 |
| Total Word Count in Article | 2772 | | |

## Summary Results

From the results summary table, we can observe some obvious patterns. Each article requires custom cutoff and adaptive filter settings. This tuning makes it possible, in most cases, to filter out large quantities of comments while maintaining a superior overall Karma point average. Out of the 10 articles, only one failed to find a good threshold point to filter out comments and bettering the overall Karma point average.

In the following section we will look at each article and analyze the results.

Article 1: Reddit.com post: Warning: Microsoft Signature PC program now requires that you can't run Linux. Lenovo's recent Ultrabooks among affected systems

This thread has a total of 438 comments, the benchmark score of the story is 1.94. and a cutoff filter is set to 5% (meaning the comment needs to score higher than 0.097 to be considered on topic). The adaptive threshold is set to 10%.

Overall, with the parameters tuned as shown previously, the adaptive filter rejected over 86% of the comments while bettering the overall karma average by 7%.

Raw Comments
- Overall karma average: 3619.05
- Overall total comments: 438

Results Without Adaptive Filter
- Good comments: 42
- Average good karma: 4248.24
- Bad comments: 396
- Average bad karma: 3552.32
- Original master word set count: 336
- Master word set count: 336

Results with Adaptive Filter
- Good comments: 60
- Average good karma: 3876.0
- Bad comments: 378
- Average bad karma: 3578.27
- Original master word set count: 336
- Master word set count: 766

Sample Comments

*Accepted Comments*
- "There's a difference between explicitly locking out linux, and using a piece of hardware that doesn't have linux driver.",
- "Yes, but it's a setting that causes the hardware to behave differently, and which would need driver support to function."

*Rejected Comments*
- "Well, I will never buy one of those. That's for sure."
- "What is this gibberish about being "on the wrong side"? Please don't make vacuous statements."

Having to read through 438 comments is a big task, the filtering pipeline managed to reduce that task down to an eighth of its original size. Just using the original filter, the system accepted 40 messages and with a larger karma mean. Using the pipeline with both filters allowed 60 messages with a slightly lower karma mean but still above the karma mean of the rejected comments. Of course, this is just one setting and can be varied for more or less comments depending on a user's needs by tweaking the thresholds.

Article 2: Kalzumeus newsletter: An A/B Testing Story

This thread has a total of 47 comments, the benchmark score of the story is 1.89 and a cutoff filter is set to 1% (meaning the comment needs to score higher than 0.0189 to be considered on topic). The adaptive threshold is set to 5%.

Overall, with the parameters tuned as shown previously, the adaptive filter rejected over 47% of the comments while bettering the overall karma average by 36%. This article happens to be one of the largest articles in the study while containing the smallest set of comments. This leads us to believe that articles with word-sets that successfully capture the topic at hand will be more successful at filtering comments from the onset. In such cases, one could forgo the adaptive filter and rely solely on the first filter especially when the pool of comments is extremely large or the comments are originating from a high-traffic site.

### Raw Comments
- Overall karma average: 8531.51
- Overall total comments: 47

### Results Without Adaptive Filter
- Good comments: 24
- Average good karma: 13826.12
- Bad comments: 23
- Average bad karma: 3006.69
- Original master word set count: 673
- Master word set count: 673

### Results with Adaptive Filter
- Good comments: 25
- Average good karma: 13294.08
- Bad comments: 22
- Average bad karma: 3119.5
- Original master word set count: 673
- Master word set count: 870

### Sample Comments

*Accepted Comments*
- "Another big aspect of the management buy-in pieces is making sure they're happy with the underlying outcome of testing: there'll be a loser. As in, it's quite difficult to get them comfortable with the notion that they've missed out on revenue/users as a result of a test, even if the winner will provide outsized gains."
- "Don't try to do this by visual pattern recognition.  Do math.  There are plenty of statistical tests that you can use, use them.  Any of them is better than looking at a graph and guessing from the shape. If you want to try to understand what is going on, learn the Central Limit Theorem.  That will let you know how fast the convergence is to the laws of large numbers.  (There are two, the strong and the weak.)

*Rejected Comments*
- "Let me guess, your company hired a "growth hacker?""
- "Perhaps you saw a slightly different version of the article, which didn't get as much click-through and so was replaced with this one."

The mean karma points of the accepted comments are much higher than the mean of the rejected comments while the adaptive filter only managed to capture one extra comment. One takeaway here is the fact that rejected messages tend to be shorter than accepted ones. This is to be expected as bigger bodies of text will have more intersecting topic words and should, in most cases, be more interesting than shorter ones.

Article 3: Science Bulletin: Researchers teleport particle of light six kilometers

This thread has a total of 154 comments, the benchmark score of the story is 1.69. and a cutoff filter is set to 1% (meaning the comment needs to score higher than 0.0169 to be considered on topic). The adaptive threshold is set to 5%.

Overall, with the parameters tuned as shown previously, the adaptive filter rejected over 15% of the comments while bettering the overall karma average by 9%.

### Raw Comments
- Overall karma average: 2882.1
- Overall total comments: 154

### Results Without Adaptive Filter
- Good comments: 110
- Average good karma: 2521.11
- Bad comments: 44
- Average bad karma: 3784.59
- Original master word set count: 303
- Master word set count: 303

### Results with Adaptive Filter
- Good comments: 131
- Average good karma: 3151.85
- Bad comments: 23
- Average bad karma: 1345.69
- Original master word set count: 303
- Master word set count: 1549

### Sample Comments

#### Accepted Comments
- "Particles are constantly randomly changing their state. Entangled particles change in tandem. As the GP said - to find out whether a particular change of state was random or a signal, one would need to compare readings from both. Readings cannot be sent faster than the speed of light. Thus preserving causality."
- "While I don't have an explanation for you, you aren't the only one to think it's weird. Einstein called it "spooky action at a distance""

*Rejected Comments*
- "10 picoseconds is "one millionth of one millionth of a second"? Damn it, I must have been misunderstanding engineering units all my life."
- "Meanwhile, in the 8th dimension, the Red Lectroids are planning their escape."

In this thread, we notice that the first filter fails to beat both the overall mean karma score and the rejected mean karma score. It is only when both filters are used together that the mean karma score of the accepted comments beats the overall and rejected scores.

## Article 4: BBC article: A 16-year-old British girl earns £48,000 helping Chinese people name their babies

This thread has a total of 199 comments, the benchmark score of the story is 1.62. and a cutoff filter is set to 1% (meaning the comment needs to score higher than 0.0162 to be considered on topic). The adaptive threshold is set to 5%.

Overall, with the parameters tuned as shown previously, the adaptive filter rejected over 11% of the comments while bettering the overall karma average by 7%. An interesting observation is the size of the master word collection. It started with 191 words from the original article and ended up with 1698, almost a 9-fold increase.

*Raw Comments*
- Overall karma average: 3666.43

- Overall total comments: 199

### Results Without Adaptive Filter
- Good comments: 150
- Average good karma: 4290.18
- Bad comments: 49
- Average bad karma: 1756.98
- Original master word set count: 191
- Master word set count: 191

### Results with Adaptive Filter
- Good comments: 178
- Average good karma: 3929.03
- Bad comments: 21
- Average bad karma: 1440.57
- Original master word set count: 191
- Master word set count: 1698

### Sample Comments

#### Accepted Comments
- "Add to that the fact that the Chinese love to substitute similar sounding words to make puns, and it's completely impossible to come up with a name that doesn't admit some negative reading."
- "I think bringing up marketing-as-information-dissemination is a red herring, because it's only a small fraction of the whole thing. The cost of telling people about a product/service should obviously be paid by the product/service provider - but we've solved that thing with a printing press, in the form of yellow pages and newspaper classifieds. Google's costs per company indexed are probably even cheaper. Marketing is by and large about gaming the way people select goods and services, so that you can sell more regardless of the quality of your offerings, or of whether the client actually needs it.",
- "Doesn't matter if it is long or short, as long as people understand it.",

#### Rejected Comments
- "Go for it! 1. You can find good translators on UpWork 2. Use Alibaba Cloud (www.aliyun.com) to setup a server anywhere in China."
- "It is close enough pronunciation wise. Downward inflection is stronger in?."

Article 5: The Bizarre Role Reversal of Apple and Microsoft

This thread has a total of 176 comments, the benchmark score of the story is 1.42 and a cutoff filter is set to 3% (meaning the comment needs to score higher than 0.0426 to be considered on topic). The adaptive threshold is set to 8%. Results without the adaptive filter fail to beat the overall karma average while the results with the adaptive filter successfully beat it.

Overall, with the parameters tuned as shown previously, the adaptive filter rejected over 40% of the comments while bettering the overall karma average by 17%.

Raw Comments
- Overall karma average: 4809.57
- Overall total comments: 176

Results Without Adaptive Filter
- Good comments: 59
- Average good karma: 2782.32
- Bad comments: 117
- Average bad karma: 5831.86
- Original master word set count: 321
- Master word set count: 321

Results with Adaptive Filter
- Good comments: 107
- Average good karma: 5769.93
- Bad comments: 69
- Average bad karma: 3320.31
- Original master word set count: 321
- Master word set count: 1537

## Sample Comments

### Accepted Comments

- "Why was the transition to HiDPI so much less painful on macOS, then? Because Apple designed the core OS to handle it much better, and make it easier to support for developers. (which was years in the making, far before the launch of rMBP.)"
- "This one was advertised as for developers/coders/engineers/designers, but I'm in a second-tier (for developers) Canadian city. Many talented tech folks move to first-tier Canadian cities, or to first tier U.S. cities, which are zeroth tier by Canadian standards, if they aren't tied down."
- "What's funny is that supporting high DPI on a winforms app isn't even hard.  Even if you size all your widgets in pixels, if you just do a little playing with the scaling options and just try out the UI on high DPI, it seems to work pretty well except for the old icons.  I forgot what the setting is called, but there's a (non-default) scaling option or two that seemed to "just work" as it were. The harder part is to consistently fix everything that uses the old win2000 8pt default font instead of a modern one."

### Rejected Comments

- "Obligatory Penny Arcade - <a href="https://www.penny-arcade.com/S=0/comic/2002/07/22" rel="nofollow">https://www.penny-arcade.com/S=0/comic/2002/07/22</a>"
- "Heh heh heh... that'll show the author."
- ""Hacker" as in "Hacker News" is from an even more distant past. The word just got appropriately decaffeinated, so as to be acceptable within our current iteration of group think."

## Article 6: Of course smart homes are targets for hackers

This thread has a total of 59 comments, the benchmark score of the story is 1.7 and a cutoff filter is set to 3% (meaning the comment needs to score higher than 0.051 to be considered on topic). The adaptive threshold is set to 8%.

Overall, with the parameters tuned as shown previously, the adaptive filter rejected over 32% of the comments but failed bettering the overall karma average.

### Raw Comments

- Overall karma average: 3687.2

- Overall total comments: 59

### Results Without Adaptive Filter
- Good comments: 26
- Average good karma: 3311.61
- Bad comments: 33
- Average bad karma: 3983.12
- Original master word set count: 206
- Master word set count: 206

### Results with Adaptive Filter
- Good comments: 40
- Average good karma: 3607.77
- Bad comments: 19
- Average bad karma: 3854.42
- Original master word set count: 206
- Master word set count: 800

### Sample Comments

#### *Accepted Comments*
- "If someone hacks my light switch all they can do is turn the light off That depends. If your light switch runs an out of date stack, it can be compromised and used as a beachhead to attack other things on your network or to run whatever software they want."
- "I looked into electronically controlling the lights in my house 15 years ago. I was hard pressed to see any value in it. Walk into a room, flick it on. Walk out, flick it off. The light is for the person in the room - why flick it on and off when nobody is there? (Yes, I know about deterring burglars)."

#### *Rejected Comments*
- "The author is alluding to taking a person home after a bar."
- "Oh god, it's going to have to be on the structural reports. "House is at risk from subsidence and there's heartbleed in the heating system""

## Article 7: Soylent halts sales of its powder as customers keep getting sick

This thread has a total of 1001 comments, the benchmark score of the story is 1.3 and a cutoff filter is set to 2% (meaning the comment needs to score higher than 0.026 to be considered on topic). The adaptive threshold is set to 10%.

The initial filter did reject almost 50% of the comments but failed to beat the overall karma average. Overall, with the parameters tuned as shown above, the adaptive filter rejected over 10% of the comments and bettered the overall karma average by 2%. This is the article that attracted the most comments in this study. It is in these types of situation that a filtering algorithm can be extremely useful to a user by shrinking the quantity of comments to more manageable sizes.

Depending on the goal of the algorithm user, if they want an even smaller set, they can increase the threshold to a more conservative setting. This will reject even more comments as long as they are okay with a lower karma average.

An other interesting observation is the size of the master word collection. It started with only 161 words from the original article and ended up with 5318, almost a 33-fold increase. This undoubtedly has to do with the huge amount of comments attached to the article. Yet, all those words didn't do much at improving the karma score. It looks like a more conservative setting on the adaptive filter may be worth experimenting with.

Raw Comments
- Overall karma average: 3289.46

- Overall total comments: 1001

### Results Without Adaptive Filter
- Good comments: 542
- Average good karma: 3518.39
- Bad comments: 459
- Average bad karma: 3019.14
- Original master word set count: 161
- Master word set count: 161

### Results with Adaptive Filter
- Good comments: 893
- Average good karma: 3370.19
- Bad comments: 108
- Average bad karma: 2621.95
- Original master word set count: 161
- Master word set count: 5318

### Sample Comments

#### *Accepted Comments*
- "Never said I consider a Soylent only diet proper. I have no opinion on that. Milk is in our diet for several thousand years, not millions. 1. There's also a lot of people who ate balanced meals and kept weight. Drinking that much milk has some unwanted stuff in it - like IGF-I - which can cause cancer in those huge amounts. 2. Maybe soy, but that's a side-effect of the dairy that uses it not for the sake of human consumption. But it's highly likely the soy used is the one for human consumption, which would remove the necessary subsidies."
- "Cheaper food? Are you serious? Soylent is not cheap, not at all. It's much more expensive than the average diet, let alone low-cost options."

#### *Rejected Comments*
- "MMM, delicious natural snake venom"
- "You seriously can't dustinguish between his writing and that of in the New Yorker?"
- "Please please go bankrupt and disappear. Soylent's whole concept is against reality, social eating and a whole food diet."

## Article 8: Google AI invents its own cryptographic algorithm

This thread has a total of 162 comments, the benchmark score of the story is 1.67 and a cutoff filter is set to 4% (meaning the comment needs to score higher than 0.0668 to be considered on topic). The adaptive threshold is set to 10%.

Overall, with the parameters tuned as shown previously, the adaptive filter rejected over 63% of the comments and bettered the overall karma average by 34%. This is quite an improvement over the overall karma average.

### Raw Comments
- Overall karma average: 2801.67
- Overall total comments: 162

### Results Without Adaptive Filter
- Good comments: 46
- Average good karma: 4047.45
- Bad comments: 116
- Average bad karma: 2307.65
- Original master word set count: 231
- Master word set count: 231

### Results with Adaptive Filter
- Good comments: 60
- Average good karma: 4229.93
- Bad comments: 102
- Average bad karma: 1961.51
- Original master word set count: 231
- Master word set count: 941

### Sample Comments

*Accepted Comments*
- "If we assume that a general AI can "understand" things in general and can "learn" over time, there is nothing stopping it from understanding the instructions it consists of, and subsequently learning how it can dynamically reprogram itself. If we extend that further, the program could also potentially obfuscate its activity by detecting logging or debugging activity. That's the way I think about it at least."
- ""The researchers didn't perform an exhaustive analysis of the encryption methods devised by Alice and Bob, but for one specific training run they observed that it was both key- and plaintext-dependent. "However, it is not simply XOR." I think this says it all."

*Rejected Comments*
- "Could have very plausibly been so, but actually I was just watching through a playlist of \xe2\x80\x9cMurderous AI\xe2\x80\x9d-trope films. (Live in Italy, Westworld not released here yet.)"
- "We can know how , but we may not know  why , if it's too complex to be understood. But yes, knowing how means that anything the computer can decode, we can too. Until the machines block access to their own source code..."

## Article 9: General questions about the Airbnb Community Commitment

This thread has a total of 160 comments, the benchmark score of the story is 1.66 and a cutoff filter is set to 4% (meaning the comment needs to score higher than 0.0664 to be considered on topic). The adaptive threshold is set to 10%.

Overall, with the parameters tuned as shown above, the adaptive filter rejected over 21% of the comments and bettered the overall karma average by 3%. This article was also the smallest of all articles in this study.

Raw Comments
- Overall karma average: 14671.23
- Overall total comments: 160

## Results Without Adaptive Filter
- Good comments: 45
- Average good karma: 19049.2
- Bad comments: 115
- Average bad karma: 12958.12
- Original master word set count: 109
- Master word set count: 109

## Results with Adaptive Filter
- Good comments: 126
- Average good karma: 15173.18
- Bad comments: 34
- Average bad karma: 12811.08
- Original master word set count: 109
- Master word set count: 1480

## Sample Comments

### Accepted Comments
- "A careful reading of your phrasing suggests that you're saying that the only people who are refused service from an Airbnb host are also in the category of people whom discrimination laws protect."
- "Airbnb actually  is  the merchant of record for all of their credit card transactions."

### Rejected Comments
- "I used to work in a camping site in Northern Italy many years ago. We were openly told not to accept people from the south of Italy as they statistically were more noisy. I didn't like it but it actually made sense. It just wasn't viable to mix Germans and Napoleteans in a tight space. I guess Germans earned that right with politness. Then one day I let gipsies in and I almost got fired."
- ""hard-left progressive" I'm not saying AirBnB has moved that much, but maybe you mean regressive left, though"

## Article 10: Cognitive bias cheat sheet

This thread has a total of 139 comments, the benchmark score of the story is 2.1 and a cutoff filter is set to 0.5% (meaning the comment needs to score higher than 0.0105 to be considered on topic). The adaptive threshold is set to 2%.

Overall, with the parameters tuned as shown previously, the adaptive filter rejected over 37% of the comments and bettered the overall karma average by 5%. This article is the largest article used in the study and may also account for the large amount of rejected comments. Unlike 'Story 2: An AB Testing Story', the other larger article in this study, the filtering here didn't fare as well in terms of karma advantage.

### Raw Comments
- Overall karma average: 1971.54
- Overall total comments: 139

### Results Without Adaptive Filter
- Good comments: 78
- Average good karma: 1946.37
- Bad comments: 61
- Average bad karma: 2003.73
- Original master word set count: 862
- Master word set count: 862

### Results with Adaptive Filter
- Good comments: 87
- Average good karma: 1971.54
- Bad comments: 52
- Average bad karma: 1791.13
- Original master word set count: 862
- Master word set count: 1331

## Sample Comments

### Accepted Comments
- "Yet polls work most of the time. I lost money when Bernie won Michigan in a surprise upset, but it told me to respect polling error. But I won money on the vast majority of my bets based on polling aggregate."
- "Isn't this the old meaning of "put your money where your mouth is"? What other meaning does it have?"

### Rejected Comments
- "Far too often, to be sure.  Two things I have yet to encounter in my (admittedly short, so far) professional career are a sane build system and version control that isn't a PITA..."
- "No, you're not a bad parent, your son is just going to grow up to be a Sith Lord."

# Differing Filter Outcomes Based on the Literary Topology of Threads

By taking a deeper look at the filtering outcomes, we can extract rough rules as to where this approach thrives and where it struggles and devise ways of finding optimum tuning parameters for both cases.

Article 7 and 9 are small in size and the adaptive filter word count soared compared to the initial filter (article 7 had a 33-fold size increase and article 9 had a 13-fold). The adaptive filter in both of those articles didn't really improve or degrade the overall karma average nor did it manage to reduce the comments by more than 20% of the total size. Article 7 had over a thousand comments while article 9 only 160, interestingly, the initial filter did a great job on article 9, but failed on article 7. We can conclude then that a smaller article may only need the initial filter when the comment pool is small. For article 7, where the comment pool was huge, a much more conservative adaptive threshold may be needed, even if it incurs a negative karma cost. This filtering approach has two goals, to prune low-quality comments and reduce the quantity of comments. A conservative tuning setting on the adaptive filter may not improve the overall karma average, but if it manages to drastically reduce the body of comments, it will have successfully reached one of its goals and may still fulfilled a user's needs.

Articles 3, 4, and 5, were all medium sized, and the adaptive filter saw a 5 to 9-fold increase in topic words compared to the initial filter. The adaptive filter in those 3 articles managed to improve the karma average while reducing the comment count. What is noteworthy here is the tuning settings of the filters. In all three articles, the difference between the initial filter threshold and the adaptive one was, on average, over 3 times larger. This may indicate that the size of an article may determine what is the best threshold ratio between filters. As size will vary from one user to the next and from one experiment to the next, it may be helpful to test different thresholds on a couple averaged-sized articles to obtain a global threshold parameter that may work on subsequent articles with minimal testing as long as they're of similar size.

## Conclusions

By using two algorithms instead of just one, we not only allow a rapid start with little pre-processing time but allow the second algorithm to focus on the evolving nature of the conversation by capturing new words. This method is both economical and accurate as shown by tracking karma points.

It is also apparent that the efficacy of this filtering pipeline varies from article to article and comment-set to comment-set. In the case of 'Article 2: Kalzumeus newsletter: An A/B Testing Story' where the article is copious and the comment set small, the filtering pipeline did a phenomenal job at reducing the quantity of comments and increasing the karma average. This is likely due to the large amount of vocabulary words available from the onset of the pipeline run. On the flip side, in the case of 'Article 6: Of course smart homes are targets for hackers' neither the first or adaptive filter managed to improve the overall karma mean score. This is probably due to the fact that the article is amongst the smaller ones in this study.

Validating if a comment is of interest or not is a very subjective task and will vary from reader to reader. Even though we were able to rely on Karma points on Hacker News as an objective, quantitative measure, other sites may not benefit from such a readily available metric. There are definitely other ways to measure the quality of a commentator. It may require a little more work, like looking at the number of followers, number of posts, frequency of posting, etc., but one should be able to emulate some sort of Karma ranking system.

In this age of overwhelming amounts of information on the Internet, supporting tools that can cut through noise are very relevant and will only become more so in the future. By experimenting with adaptive natural language filtering systems, we can attempt to find value in crowd-sourced material at world scale – the practical applications of such tools are exciting and far reaching.

Here are a few ways this study can be ameliorated to extend accuracy and purpose:

- This study used only 10 articles to test, analyze and extrapolate conclusions. An interesting next step would be to increase the number of articles to hundreds or even thousands to measure the statistical effect and significance this approach has on karma scores.
- Grading a message thread and adjusting frequencies based on how quickly it degrades.
- Re-filtering the first comments so they can benefit from a more educated adaptive filter.
- Scoring commentators over many comments to insure that they don't get rejected if they are, on average, a good participant.
- Focus the tool on low-end comments instead of high-end ones to function as a spam and automated comment filter. This could entail using a global list of spam terms on top of an understanding of the thread's terminology and context.
- Using tools like wordv2vec mapping to find all allowed and related words and not having to use adaptive filters.
- Using tools such as words2map that finds word-vector representation boosted with Internet terms to capture a more universal understanding beyond the original text.

# Appendix

## Filtering Pipeline Algorithm (Python source code)

```python
# ----------------------------------
# libraries & functions
# ----------------------------------
import time
import nltk
import string
import pandas as pd
import numpy as np

def prep_text_to_sentences(text_to_prep):
    # clean raw text - eliminate special characters, numbers, and force to lower case
    import string, re
    regex = re.compile('[^a-z]')
    text_to_prep = text_to_prep.replace('.', 'ootoo').replace(';', 'ootoo').replace('?', 'ootoo').replace('!', 'ootoo').replace('\n', 'ootoo').lower()
    text_to_prep = regex.sub(' ', text_to_prep)
    text_to_prep = re.sub(' +',' ',text_to_prep)
    text_to_prep = text_to_prep.split('ootoo')
    text_to_prep = [x for x in text_to_prep if x.strip()]
    return text_to_prep

def manage_new_words(content, current_words, add_new_words_to_master=False, min_word_length=4):
    # clean raw text - eliminate special characters, and force to lower case
    content = prep_text_to_sentences(content)

    # tokenize corpus
    words = nltk.word_tokenize(" ".join(content))

    # keep acronyms
    # if word is between 2 and min_word_length but all CAPS then keep
    acro_words = [word for word in words if len(word) > 1 if len(word) < min_word_length if word.isupper()]

    # Remove words smaller than min_word_length
    words = [word for word in words if len(word) >= min_word_length]

    words = words + acro_words
    # append any existing list of words to new set of words
    if (add_new_words_to_master):
        words += current_words
        words = list(set(words))
    else:
        words = current_words

    # remove none reserved words
    clean_content = [' '.join(w for w in sentence.split() if w.lower() in words) for sentence in content]

    # assumption - removing blanks?
    clean_content = [note for note in clean_content if note.strip() != '']

    word_count_in_sentences = [len(w.split()) for w in clean_content]

    return clean_content, words, word_count_in_sentences;
```

```python
# -----------------------------------
# user paramters
# -----------------------------------
# minimum mean entry term frequency required to qualify as good comment
USER_DEFINED_MIN_RELATEDNESS_PERCENT = 0.05
# minimum mean entry term frequency required to have content added to master-word list
USER_DEFINED_BEST_RELATEDNESS_PERCENT = 0.1
# turn on or off the ability to add new words to the master-word list
GROW_MASTER_WORDS = True

# -----------------------------------
# test articles
# -----------------------------------

# 4 articles available to test
article_number = 10

if (article_number==1):
        # https://www.reddit.com/r/linux/comments/53ri0m/warning_microsoft_signature_pc_program_now/
        # std  0.05 , growth  0.1
        USER_DEFINED_MIN_RELATEDNESS_PERCENT = 0.05
        USER_DEFINED_BEST_RELATEDNESS_PERCENT = 0.1
        file = "Warning Microsoft Signature PC program now requires that you cant run Linux.txt"
        all_comments_df = pd.read_csv('Warning Microsoft Signature PC program now requires that you cant run Linux_full_comments.csv')

elif (article_number==2):
        # https://training.kalzumeus.com/newsletters/archive/ab-testing-story
        # std  0.01 , growth  0.05
        USER_DEFINED_MIN_RELATEDNESS_PERCENT = 0.01
        USER_DEFINED_BEST_RELATEDNESS_PERCENT = 0.05
        file = "AnABTestingStory.txt"
        all_comments_df = pd.read_csv('AnABTestingStory_full_comments.csv')

elif (article_number==3):
        # http://sciencebulletin.org/archives/5445.html
        # std  0.01 , growth  0.05
        USER_DEFINED_MIN_RELATEDNESS_PERCENT = 0.01
        USER_DEFINED_BEST_RELATEDNESS_PERCENT = 0.05
        file = "Researchers teleport particle of light six kilomet.txt"
        all_comments_df = pd.read_csv('Researchers teleport particle of light six kilometres_full_comments.csv')

elif (article_number==4):
        # http://www.bbc.co.uk/newsbeat/article/37255033/a-16-year-old-british-girl-earns-48000-helping-chinese-people-name-their-babies
        # std  0.01 , growth  0.05
        USER_DEFINED_MIN_RELATEDNESS_PERCENT = 0.01
        USER_DEFINED_BEST_RELATEDNESS_PERCENT = 0.05
        file = "A 16-year-old British girl earns £48,000 helping C.txt"
        all_comments_df = pd.read_csv('A 16-year-old British girl earns £48,000 helping C_full_comments.csv')

elif (article_number==5):
        # https://backchannel.com/the-bizarre-role-reversal-of-apple-and-microsoft-25d8b391d5b0#.mc73k791l
        # std 0.03, growth 0.08
```

```python
            USER_DEFINED_MIN_RELATEDNESS_PERCENT = 0.03
            USER_DEFINED_BEST_RELATEDNESS_PERCENT = 0.08
            file = "The Bizarre Role Reversal of Apple and Microsoft.txt"
            all_comments_df = pd.read_csv('The Bizarre Role Reversal of Apple and Microsoft_full_comments.csv')

    elif (article_number==6):
            # https://mjg59.dreamwidth.org/45483.html
            # std 0.03, growth 0.08
            USER_DEFINED_MIN_RELATEDNESS_PERCENT = 0.03
            USER_DEFINED_BEST_RELATEDNESS_PERCENT = 0.08
            file = "Of course smart homes are targets for hackers.txt"
            all_comments_df = pd.read_csv('Of course smart homes are targets for hackers_full_comments.csv')

    elif (article_number==7):
            # http://www.latimes.com/business/technology/la-fi-tn-soylent-recall-20161027-story.html
            # std 0.02 , growth 0.1
            USER_DEFINED_MIN_RELATEDNESS_PERCENT = 0.02
            USER_DEFINED_BEST_RELATEDNESS_PERCENT = 0.1
            file = "Soylent halts sales of its powder as customers kee.txt"
            all_comments_df = pd.read_csv('Soylent halts sales of its powder as customers keep getting sick_full_comments.csv')

    elif (article_number==8):
            # http://arstechnica.co.uk/information-technology/2016/10/google-ai-neural-network-cryptography/
            # std 0.04 , growth  0.1
            USER_DEFINED_MIN_RELATEDNESS_PERCENT = 0.04
            USER_DEFINED_BEST_RELATEDNESS_PERCENT = 0.1
            file = "Google AI invents its own cryptographic algorithm.txt"
            all_comments_df = pd.read_csv('Google AI invents its own cryptographic algorithm_full_comments.csv')

    elif (article_number==9):
            # https://www.airbnb.com/help/article/1523/general-questions-about-the-airbnb-community-commitment?topic=533
            # std  0.04 , growth 0.1
            USER_DEFINED_MIN_RELATEDNESS_PERCENT = 0.04
            USER_DEFINED_BEST_RELATEDNESS_PERCENT = 0.1
            file = "General questions about the Airbnb Community Commi.txt"
            all_comments_df = pd.read_csv('General questions about the Airbnb Community Commitment_full_comments.csv')

    elif (article_number==10):
            # https://betterhumans.coach.me/cognitive-bias-cheat-sheet-55a472476b18#.lujwtbglx
            # std  0.005 , growth  0.02
            USER_DEFINED_MIN_RELATEDNESS_PERCENT = 0.005
            USER_DEFINED_BEST_RELATEDNESS_PERCENT = 0.02
            file = "Cognitive bias cheat sheet.txt"
            all_comments_df = pd.read_csv('Cognitive bias cheat sheet_full_comments.csv')

# ----------------------------------
# load and process article
# ----------------------------------

with open(file, 'r') as myfile:
    raw_article=myfile.read()

master_word_set = []
article, master_word_set, word_count_in_sentences = manage_new_words(raw_article, master_word_set, add_new_words_to_master=True)
```

```python
sentence_count = np.array([len(t.split()) for t in (' '.join(w for w in comment.split() if w.lower() in master_word_set) for comment in article)])
article_overall_mean_score = np.mean([t/float(len(master_word_set)) for t in sentence_count]) * len(article)

# ------------------------------------
# stream (loop) process and evaluate
# comments
# ------------------------------------

counter = -1
good_comments = {}
bad_comments = {}
overall_scores = []
good_karma = []
bad_karma = []
all_karma = []
adapting_master_word_set = master_word_set
for index, row in all_comments_df.iterrows():
        if (isinstance(row['comment'], basestring)==False):
                continue

        counter += 1
        print(counter)
        comment_raw, adapting_master_word_set, sentence_count = manage_new_words(row['comment'], adapting_master_word_set)
        # if counter > 3: break

        if  (len(sentence_count) > 0):
                comment_score = np.mean([t/float(len(master_word_set)) for t in sentence_count]) * len(comment_raw)

                print('overall score:', comment_score, 'karma', row['karma'])
                overall_scores.append(comment_score)

                if (GROW_MASTER_WORDS==True) & (comment_score >= (
                        USER_DEFINED_BEST_RELATEDNESS_PERCENT * article_overall_mean_score)):
                                print('growing master word list')
                                comment_raw, adapting_master_word_set, sentence_count = manage_new_words(row['comment'], adapting_master_word_set, add_new_words_to_master=True)

                if comment_score > (USER_DEFINED_MIN_RELATEDNESS_PERCENT * article_overall_mean_score):
                        good_comments[counter] = row['comment']
                        good_karma.append(row['karma'])
                else:
                        bad_comments[counter] = row['comment']
                        bad_karma.append(row['karma'])
        else:
                overall_scores.append(0)
                bad_comments[counter] = row['comment']
                bad_karma.append(row['karma'])
        all_karma.append(row['karma'])

print('---------------------------------------')
print('Second filter:', GROW_MASTER_WORDS)
print('Minimum cutoff for standard filter:', USER_DEFINED_MIN_RELATEDNESS_PERCENT)
print('Minimum cutoff for related content:', USER_DEFINED_BEST_RELATEDNESS_PERCENT)
print('Article score:', article_overall_mean_score)
print('Overall Average Karma:', np.mean(all_karma))
```

```
print('Good comments:',len(good_comments))
print('Average good karma:',np.mean(good_karma))
print('Bad comments:',len(bad_comments))
print('Average bad karma:',np.mean(bad_karma))

print('Original master word set count', len(master_word_set))
print('New master word set count', len(adapting_master_word_set))

print('Overall success rate:', (np.mean(good_karma)-np.mean(all_karma)) /  np.mean(good_karma))
print('Overall rejection comments:',  float(len(good_comments)) / (len(good_comments) + len(bad_comments)) - 1)
print('Total comments:', len(good_comments)+len(bad_comments))
```

## Hacker News Comment Gatherer (Python source code)

```
import time
import requests
import json
import html
import HTMLParser
import nltk

file_name = 'Cognitive bias cheat sheet'
def prep_text_to_sentences(text_to_prep):
    import string, re
    regex = re.compile('[^a-z]')
    text_to_prep = text_to_prep.replace('.', 'ootoo').replace(';', 'ootoo').replace('?', 'ootoo').replace('!', 'ootoo').replace('\n', 'ootoo').lower()
    text_to_prep = regex.sub(' ', text_to_prep)
    text_to_prep = text_to_prep.split('ootoo')
    return text_to_prep;

def get_comments(comment_id):
    entry = requests.get("https://hacker-news.firebaseio.com/v0/item/" + str(comment_id) + ".json?print=pretty")
    text = ''
    new_kids = None
    karma_points = None
    if (entry.json()==None):
        return None, None
    if ('deleted' in entry.json().keys()):
        return None, None, None
    if ('text' in entry.json().keys()):
        text = entry.json()['text']
    if ('by' in entry.json().keys()):
        # get karma points
        # https://news.ycombinator.com/user?id=trebor
        user_info = requests.get("https://hacker-news.firebaseio.com/v0/user/"+ str(entry.json()['by']) + ".json?print=pretty")
        if ('karma' in user_info.json().keys()):
            karma_points = user_info.json()['karma']
    if ('kids' in entry.json().keys()):
        new_kids = entry.json()['kids']
    # introduce pause
    time.sleep(0.2)
    return HTMLParser.HTMLParser().unescape(text).replace('<p>',' ').replace('</i>',' ').replace('<i>',' '), new_kids, karma_points;
    # return html.unescape(text).replace('<p>',' ').replace('</i>',' ').replace('<i>',' '), new_kids, karma_points;

def get_all_comments_for_story(story_id):
    import pandas as pd
    entry = requests.get("https://hacker-news.firebaseio.com/v0/item/" + str(story_id) + ".json?print=pretty")
```

```python
        comments = []
        karmas = []
        if ('title' in entry.json().keys()):
            comments.append(entry.json()['title'])
        if ('text' in entry.json().keys()):
            comments.append(entry.json()['text'])
        karmas.append(-1)

        if ('kids' in entry.json().keys()):
            kids_temp =  entry.json()['kids']

            while len(kids_temp) > 0:
                current_kids = kids_temp
                kids_temp = []
                for kid_id in current_kids:
                    print('collecting ID:' + str(kid_id))
                    comment, kids, karma = get_comments(kid_id)
                    if (comment is not None):
                        comments.append(comment)
                    if (karma is not None):
                        karmas.append(karma)
                    if (kids is not None):
                        for kid in kids:
                            kids_temp.append(kid)

    comment_data = pd.DataFrame({'comment' : comments, 'karma' : karmas})
    return (comment_data);

# get story by hacker news id
comment_data = get_all_comments_for_story(12804870)
print(len(comment_data))
comment_data.to_csv(file_name + "_full_comments.csv", index=False, encoding='utf-8')
```